\documentclass[11pt]{article} % For LaTeX2e
\usepackage[preprint]{tmlr}

\usepackage{amsthm,amsmath,amssymb,bbm}

\usepackage[colorlinks,citecolor=blue,urlcolor=blue]{hyperref}
\usepackage{url}

\theoremstyle{plain}
\newtheorem{theorem}{Theorem}
\newtheorem{assume}{Assumption}
\newtheorem{lemma}{Lemma}
\newtheorem{cor}{Corollary}

\newtheorem{exm}{Example}
\newtheorem{remark}{Remark}

\allowdisplaybreaks

\title{Misclassification excess risk bounds for PAC-Bayesian classification via convexified loss}
\author{\name The Tien Mai \email the.t.mai@ntnu.no 
	\\
	\addr Department of Mathematical Sciences,
	\\
	Norwegian University of Science and Technology, 
	\\
	Trondheim 7034, Norway.
}

\def\month{MM}  % Insert correct month for camera-ready version
\def\year{YYYY} % Insert correct year for camera-ready version
\def\openreview{\url{https://openreview.net/forum?id=XXXX}} % Insert correct link to OpenReview for camera-ready version

\begin{document}

\maketitle

\begin{abstract}	
PAC-Bayesian bounds have proven to be a valuable tool for deriving generalization bounds and for designing new learning algorithms in machine learning. However, it typically focus on providing generalization bounds with respect to a chosen loss function. In classification tasks, due to the non-convex nature of the 0-1 loss, a convex surrogate loss is often used, and thus current PAC-Bayesian bounds are primarily specified for this convex surrogate. This work shifts its focus to providing misclassification excess risk bounds for PAC-Bayesian classification when using a convex surrogate loss. Our key ingredient here is to leverage   PAC-Bayesian relative bounds in expectation rather than relying on PAC-Bayesian bounds in probability. We demonstrate our approach in several important applications.
\end{abstract}


\section{Introduction and motivation}

Building on the foundational works initiated by \cite{shawe1997pac,mcallester1998some,mcallester1999pac}, PAC-Bayesian theory has become a crucial framework not only for deriving generalization bounds but also for developing novel learning algorithms in machine learning \citep{catonibook,guedj2019primer,alquier2021user,rivasplata2022pac}. While PAC-Bayesian bounds traditionally address risk bounds with respect to a specific loss function, in classification tasks, the inherently non-convex and non-smooth nature of the 0-1 loss necessitates the use of a convex surrogate loss to facilitate computation \citep{zhang2004statistical,bartlett2006convexity}. Several PAC-Bayes studies have addressed this by incorporating convex surrogate losses: \cite{dalalyan2012mirror} and \cite{alquier2016properties} both focusing on risk bounds for the convexified loss. Although recent research has made strides in using PAC-Bayesian techniques to establish prediction bounds in classification, these efforts have not succeeded in providing misclassification risk bounds \citep{cottet20181,mai2023reduced,mai2023high}. This paper aims to address this gap by focusing on misclassification excess risk bounds in PAC-Bayesian classification using a convex surrogate loss.

We formally consider the following general binary classification. Given a covariate/feature $ X \in \mathcal{X} $, one has that the class label 
$ Y = 1 $ with probability  $ p(X),
$ and $ Y=-1 $ with probability $ 1 - p(X) $.
The accuracy of a classifier $\eta$
is defined by the prediction or misclassification error, given as
$$
R_{0/1} (\eta)
=
\mathbb{P} (Y \neq \eta(X)).
$$
The Bayes classifier, $ \eta^*(X)= {\rm sign} (p( X) - 1/2) $, is widely recognized for minimizing  $ R_{0/1}(\eta) $
 \citep{vapnik,devroye1997probabilistic}, i.e.
$$
R^*_{0/1} := R_{0/1} (\eta^*)
=
\inf R_{0/1} (\eta)
.
$$
With \( p(X) \) being unknown, a classifier \( \hat{\eta}(X) \) needs to be designed using the available data: a random sample of $n$ independent observations $ D_n = \{ ( x_1,y_1)$, $ \ldots, $ $ ( x_n,y_n) \} $. 
The design points $ x_i$ may be considered as fixed or random.  
The corresponding (conditional) prediction error of $\hat{\eta}$ is now as
$$
R_{0/1} (\hat{\eta})
=
\mathbb{P} (Y \neq \hat{\eta}(X) \mid D_n)
$$
and the goodness of $\hat{\eta}$  with respect to $\eta^*$ is measured by the \textit{misclassification
excess risk}, defined as 
$$
\mathbb{E} \, R_{0/1} (\hat{\eta}) - R^*_{0/1}
=
\mathbb{E} \, R_{0/1} (\hat{\eta})-R_{0/1} (\eta^*).
$$
The empirical risk minimization method is a general nonparametric approach to determine a classifier \(\hat{\eta}\) from data, where the true prediction error \( R_{0/1}(\eta) \) minimization is replaced by the minimization of the empirical risk $ r_n^{0/1} $ over a specified class of classifiers, $ \{ \eta_{\theta}: \mathcal{X}\rightarrow \{-1,1\} ,\,\theta\in\Theta\} $, where $ r_n^{0/1} $ is given by:
\begin{align*}
r_n^{0/1}(\theta) 
= 
\frac{1}{n}\sum_{i=1}^n   
\mathbbm{1} \{ y_{i} \neq \eta_{\theta}( x_i) \} 
.
\end{align*}
PAC-Bayesian classification using the 0-1 loss was thoroughly examined in a series of works by Olivier Catoni over 20 years ago, in \cite{catoni2003PAC, catoni2004statistical, catonibook}. However, due to the computational challenges posed by the non-convexity of the zero-one loss function, particularly when dealing with huge and/or high-dimensional data, a convex surrogate loss is often preferred to simplify the computational problem. The convex surrogate loss in PAC-Bayesian approach for classification has been considered in various studies. For example, \cite{alquier2016properties} explored a variational inference approach for PAC-Bayesian methods, emphasizing the importance of convexified loss, while \cite{dalalyan2012mirror} and \cite{mai2023high} investigated PAC-Bayesian classification using convex surrogate loss and gradient-based sampling methods such as Langevin Monte Carlo. 

PAC-Bayesian bounds as in  \cite{alquier2021user}, when using a convexified loss, often leads to prediction bounds or excess risk with respect to the convexified loss. To the best of our knowledge, misclassification excess risk bounds for PAC-Bayesian classification when using convexified loss have not yet been established. In this work, we provide a unified procedure to obtain such results. Our work is carried out under the so-called low-noise condition. The low-noise condition described is a common assumption in the classification literature, as seen in works such as \cite{mammen1999smooth,tsybakov2004optimal,bartlett2006convexity}. The main challenge for any classifier typically lies near the decision boundary \( \{x : p(x) = 1/2\} \). In this region, accurately predicting the class label is particularly difficult because the label information is predominantly noisy. Given this, it is reasonable to assume that \( p(x) \) is unlikely to be very close to \( 1/2 \).

In the subsequent section, Section \ref{sc_problem_method}, we introduce our primary notations and present our main results. In Section \ref{sc_application}, we apply our general procedure to two significant applications: high-dimensional sparse classification and 1-bit matrix completion. To the best of our knowledge, the results obtained for these two problems are novel. We conclude our work in Section \ref{sc_conclusion}, while all technical proofs are provided in Appendix \ref{sc_proof}.

\section{Main result}
\label{sc_problem_method}
\subsection{PAC-Bayesian framework}
\label{sectionnotation}

We observe an i.i.d sample $(X_1,Y_1),\dots,(X_n,Y_n)$, of a random pair $ (X,Y) $ taking values in $\mathcal{X}\times \{-1,1\} $,  
from the same distribution $ \mathbb{P} $. A set of classifiers is chosen by the user:
$ \{ \eta_{\theta}: \mathcal{X}\rightarrow \{-1,1\} ,\,\theta\in\Theta\} $.
For example, one may have $ \eta_{\theta}(x) = {\rm sign} (\left<\theta,x\right>) \in \{-1,1\} $. In this paper, the symbol $\mathbb{E}$ will always denote the expectation with respect to the (unknown) law $\mathbb{P} $ of the $(X_i,Y_i)$'s.

Consider a convex loss surrogate function $ \phi: \mathbb{R}^2 \rightarrow \mathbb{R}^+ $, the empirical convex risk is defined as
$$
r_n^\phi (\theta)
:=
\frac{1}{n}\sum_{i=1}^n \phi_i (\theta)
:=
\frac{1}{n}\sum_{i=1}^n \phi(Y_i, \eta_{\theta}(X_i)) ,
$$
and its expected risk is given as
$
R^\phi (\theta)
=
\mathbb{E}[\phi(Y, \eta_{\theta}(X)) ]
.
$
\\
Convex loss functions commonly used in classification include logistic loss and hinge loss. More examples can be found for example in \cite{bartlett2006convexity}.

Let \( \mathcal{P}(\Theta) \) denote the set of all probability measures on \(\Theta\). We define a prior probability measure \(\pi(\cdot)\) on the set \(\Theta\). For any \(\lambda > 0\), as in the PAC-Bayesian framework \cite{catonibook,alquier2021user}, the Gibbs posterior \(\hat{\rho}_{\lambda}^\phi\), with respect to the convex loss $ \phi, $ is defined by
	\begin{equation}
	\label{def:Gibbspost}
	\hat{\rho}_{\lambda}^\phi({\rm d}\theta) 
= 
\frac{\exp[-\lambda r_n^\phi(\theta)]}
{\int \exp[-\lambda r_n^\phi] {\rm d}\pi } \pi({\rm d}\theta)
, 	
	\end{equation}
	and our mean estimator is defined by
		$
\hat{\theta} = \int \theta	\hat{\rho}_{\lambda}^\phi({\rm d}\theta) 
	.
	$
From now, we will let $\theta^*$ denote a minimizer of $ R^\phi $ when it exists:
$
R^\phi(\theta^*) 
= 
\min_{\theta\in\Theta} R^\phi(\theta)
. 
$
\\
In PAC-Bayes theory, when utilizing a \(\phi\)-loss function, it is customary to regulate the excess \(\phi\)-risk, 
$$
R^\phi (\theta) - R^\phi (\theta^*) 
$$
see e.g. \cite{alquier2021user}. However, in classification tasks, it is equally crucial to control the misclassification excess risk, \( \mathbb{E} \, R_{0/1} (\theta) - R^*_{0/1} \), which is the primary focus of this paper.

\subsection{Main result}
\subsubsection{Assumptions}
Certain conditions are essential for deriving our main result.

\begin{assume}[Bounded loss]
	\label{assume_boundedloss}
The convex surrogate loss function \(\phi\) is assumed to be bounded, with its values lying in the range \([0, B]\).
\end{assume}

\begin{assume}[Lipschitz loss]
	\label{assume_Lipschitz}
	We assume that the loss function $ \phi (y,\cdot) $ is $ L $-Lipschitz in the sense that there exist some constant $ L>0 $ such that
	$
	|\phi (y,\eta_{\theta}(x) ) - \phi (y,\eta_{\theta'}(x))| 
	\leq 
	L \| \theta- \theta' \|
	.
	$
\end{assume}

\begin{assume}[Bernstein condition]
	\label{dfnbernstein}
Assuming that there is a constant $K>0 $ such that, for any $\theta\in\Theta$,
$
	\| \theta - \theta^* \|_2^2 
	\leq 
	K [R^\phi (\theta)-R^\phi (\theta^*) ]
	.
$
\end{assume}

\begin{assume}[Margin condition]
	\label{asume_margin}
	We assume that there exist a constant $ c>0 $ such that
	\begin{align*}
	\mathbb{P} \left\{ 0<|p(X) - 1/2|< 1/(2c) \right\} = 0
	.
	\end{align*}
\end{assume}

The boundedness condition in Assumption \ref{assume_boundedloss} is not central to our analysis; rather, it serves to simplify the presentation and enhance the clarity of the paper. It is important to note that PAC-Bayesian bounds can also be derived for unbounded loss functions, as discussed in \cite{alquier2021user}.

Assumption \ref{assume_Lipschitz} and \ref{dfnbernstein} have been extensively studied in various forms in the learning theory literature, such as \citep{mendelson2008obtaining,zhang2004statistical,alquier2019estimation,elsener2018robust,alaya2019collective}. Some examples of the loss functions that are 1-Lipschitz are: hinge loss $ \phi(y,y') = \max (0,1-yy') $ and logistic loss $ \phi(y,y') = \log (1+\exp(-yy')) $. Assumption  \ref{dfnbernstein} implicitly means that our predictors are identifiability. 

\begin{remark}
It is worth noting that our Bernstein condition in Assumption \ref{dfnbernstein} is slightly stronger than the one considered in \cite{alquier2021user}. Specifically, Definition 4.1 in \cite{alquier2021user} defines a Bernstein condition where there exists a constant \( K > 0 \) such that for any \( \theta \in \Theta \),
\[
\mathbb{E} \left\{ \left[  \phi_i(\theta) - \phi_i(\theta^*) \right]^2 \right\} \leq K [R^\phi (\theta) - R^\phi (\theta^*) ].
\]
Therefore, if we additionally assume that the loss function \( \phi \) in our context is further \( L \)-Lipschitz, then
$
\mathbb{E} \left\{ \left[  \phi_i(\theta) - \phi_i(\theta^*) \right]^2 \right\} \leq L^2 \mathbb{E} \| \theta - \theta^* \|_2^2 \leq L^2 K [R^\phi (\theta) - R^\phi (\theta^*) ],
$
which satisfies Definition 4.1 in \cite{alquier2021user}.	
\end{remark}

The low-noise condition described in Assumption \ref{asume_margin} is a common assumption in the classification literature, as seen in works such as \citep{abramovich2018high,tsybakov2004optimal,mammen1999smooth,bartlett2006convexity}. The main challenge for any classifier typically lies near the decision boundary \( \{x : p(x) = 1/2\} \), which in logistic regression corresponds to the hyperplane \( \theta^\top x = 0 \), where \( p(x) = (1 + e^{-\theta^\top x})^{-1} \). In this region, accurately predicting the class label is particularly difficult because the label information is predominantly noisy. Given this, it is reasonable to assume that \( p(x) \) is unlikely to be very close to \( 1/2 \).

\subsubsection{Main results}

While high probability PAC-Bayes bounds for the excess \(\phi\)-risk, \( R^\phi (\theta) - R^\phi (\theta^*) \), are frequently discussed in the literature (see e.g. \cite{alquier2021user}), PAC-Bayes bounds in expectation have received comparatively less attention. Utilizing high probability PAC-Bayes bounds for deriving prediction bounds has also been explored to some extent, as evidenced by several works such as \cite{cottet20181,mai2023reduced,mai2023high}. However, these approaches often do not provide bounds for misclassification excess risk unless under strictly noiseless conditions.

In this study, we illustrate the utility of PAC-Bayes bounds in expectation for deriving misclassification excess risk bounds. Specifically, we first introduce a PAC-Bayesian relative bound in expectation, which is a slight extension of Theorem 4.3 in \cite{alquier2021user}. For two probability distributions \(\mu $ and $\nu$ in $ \mathcal{P}(\Theta) \), let  \(\mathcal{K}(\nu \| \mu)\) denote the Kullback-Leibler divergence from \(\nu\) to \(\mu\). 

Put $ \overline{C} := \max(2L^2K, B)  $.

\begin{theorem}
	\label{theorembernstein}
Assuming that Assumptions \ref{assume_boundedloss}, \ref{assume_Lipschitz} and \ref{dfnbernstein} are satisfied, let's take \( \lambda = n/\overline{C}  \). Then we have:
\begin{equation*}
\mathbb{E} [ \mathbb{E}_{\theta\sim\hat{\rho}_{\lambda}^\phi} [R^\phi(\theta) ]] - R^\phi (\theta^*) 
\leq 
2 \!\!
 \inf_{\rho\in\mathcal{P}(\Theta)}  
\left\{  \mathbb{E}_{\theta\sim\rho}
 [R^\phi (\theta) ] - R^\phi (\theta^*) 
 + 
 \frac{\overline{C}  
 	\mathcal{K}(\rho\| \pi) }{n} \right\}
 .
\end{equation*}

\end{theorem}

The proof is given in Appendix \ref{sc_proof}. As discussed in \cite{catonibook,alquier2021user}, the bound in Theorem \ref{theorembernstein} can be employed to derive error rates for the excess \(\phi\)-risk in a general setting as follows: one needs to find a \(\rho_\epsilon\) such that
$
\mathbb{E}_{\theta \sim \rho_\epsilon} [R^\phi(\theta) ] 
\simeq R^\phi(\theta^*) + \frac{\epsilon}{n} 
$
and ensure that \( \mathcal{K}(\rho_\epsilon \| \pi) \simeq \epsilon \) to obtain:
$
\mathbb{E}[ \mathbb{E}_{\theta\sim\hat{\rho}_{\lambda}} [R^\phi (\theta) ]]
\lesssim 
R^\phi (\theta^*)
+ \frac{\epsilon}{n}
 + \frac{2 \overline{C}  \epsilon }{n} 
 .
$
Hence the rate is of order $1/n$.

\begin{remark}
One can derive a PAC-Bayesian relative bound without invoking the Bernstein condition from Assumption \ref{dfnbernstein}, see e.g \cite{alquier2021user}. Nevertheless, this results in a slower convergence rate of order $ n^{-1/2} $. In contrast, under the low-noise condition specified in Assumption \ref{asume_margin}, which is our primary assumption, it is well-known that a faster rate of order $ 1/n $ can be obtained \cite{abramovich2018high,tsybakov2004optimal}. Hence, the need for imposing the Bernstein condition in Assumption \ref{dfnbernstein} becomes crucial.
\end{remark}

The following theorem presents our main results on misclassification excess risk bounds for PAC-Bayesian classification approaches using convexified loss. The strategy involves utilizing a broad result from \cite{bartlett2006convexity}. To establish our main result presented in Theorem \ref{thm_main} below, we further assume that the $ \phi $-loss function is \textit{classification-calibrated}. Specifically, for 
$ 
\zeta \in [0,1], \zeta \neq 1/2,
$  
the following condition must hold:
$$ \inf_{\alpha\in\mathbb{R}} C_{\zeta} (\alpha) <  \inf_{\alpha: \alpha(2\zeta -1 ) \leq 0 } C_{\zeta} (\alpha) 
,
$$
where $ C_{\zeta} (\alpha) = \zeta \phi(\alpha) + (1-\zeta)\phi(-\alpha) $. 
This is a minimal requirement, indicating that the $\phi$-loss function possesses the same capacity for classification as the Bayes classifier. For a more detailed discussion, refer to \cite{bartlett2006convexity}.

\begin{theorem}
	\label{thm_main}
Assuming both Theorem \ref{theorembernstein} and Assumption \ref{asume_margin} hold, and by selecting \(\lambda = n / \overline{C} \), there exists a constant \(\Psi > 0\) such that
	\begin{equation}
	\label{eq_mai_01}
	\mathbb{E} [
\mathbb{E}_{\theta\sim \hat{\rho}_{\lambda}^\phi}
	[ R_{0/1} (\theta) ] ] - R^*_{0/1}
	\leq
\Psi \!\!\!
 \inf_{\rho\in\mathcal{P}(\Theta)} \! 
\left\{  \mathbb{E}_{\theta\sim\rho}
[R^\phi (\theta) ] - R^\phi (\theta^*) 
+ 
\frac{\overline{C}  
	\mathcal{K}(\rho\| \pi) }{n} \right\}
	,
	\end{equation}
	and
	\begin{equation}
	\label{eq_mai_02}
\mathbb{E} [
R_{0/1} ( \hat\theta) ] - R^*_{0/1}
\leq
\Psi \!\!\!
 \inf_{\rho\in\mathcal{P}(\Theta)}  
\left\{  \mathbb{E}_{\theta\sim\rho}
[R^\phi (\theta) ] - R^\phi (\theta^*) 
+ 
\frac{\overline{C}  
	\mathcal{K}(\rho\| \pi) }{n} \right\}
.
\end{equation}	
\end{theorem}

\begin{remark}
	\label{rm_01}
Similar to Theorem \ref{theorembernstein}, the bound in Theorem \ref{thm_main} can be utilized to derive general misclassification error rates. For instance, since the bound in \eqref{eq_mai_01} holds for any \( \rho \in \mathcal{P}(\Theta) \), one can specify a distribution \( \rho_\delta \) such that
	$ 
	\mathbb{E}_{\theta\sim\rho_\delta} [R^\phi(\theta) ] 
	- R^\phi(\theta^*) \lesssim \delta/n 
	$ 
	and that $ \mathcal{K}(\rho_\delta\| \pi) \lesssim \delta $ and consequently:
$
	\mathbb{E} [
\mathbb{E}_{\theta\sim \hat{\rho}_{\lambda}^\phi}
[ R_{0/1} (\theta) ] ] - R^*_{0/1}
	\lesssim  
\frac{\delta}{n} + \frac{2 \overline{C}  \delta }{n} ,
$
	hence the misclassification excess rate can be of the order $1/n$. Some classical examples are given below.
\end{remark}

From Theorem \ref{thm_main}, we immediately obtain the following corollary regarding the $ \ell_2 $ error for the predictor.

\begin{cor}
	\label{cor_estimation}
Assuming that Theorem \ref{thm_main} is satisfied and let's take \( \lambda = n/ \overline{C} \). Then, with some universal constant \( C > 0 \), we have that
	\begin{equation*}
	\mathbb{E}\,
	\mathbb{E}_{\theta\sim \hat\rho_\lambda^\phi}\!
\left[ 
\| \theta - \theta^* \|_2^2 \right]
\leq 
C \!\!\!
	\inf_{\rho\in\mathcal{P}(\Theta)}
	\left\{ 
	\mathbb{E}_{\theta\sim\rho}
	[ R^\phi(\theta) - R^\phi (\theta^*) ] 
	+  
	\frac{\mathcal{K}(\rho\|\pi) + \log\frac{2}{\varepsilon}}{\lambda} 
	\right\}
	.
	\end{equation*}
\end{cor}

With the same rationale as provided in Remark \ref{rm_01}, some error rates can be obtained from Corollary \ref{cor_estimation}.

\begin{remark}
It is crucial to recognize that, in the absence of Assumption \ref{asume_margin}, one may not achieve a result analogous to Theorem \ref{thm_main}. For instance, as demonstrated by \cite{zhang2004statistical}, for the logistic loss, 
$ \mathbb{E} [ R_{0/1} (\theta) ] - R^*_{0/1} \lesssim 
 (\mathbb{E} [ R^\phi(\theta)] - R^\phi (\theta^*)  )^{1/2}
$. Consequently, it is generally unlikely to derive a comparable result for PAC-Bayesian methods without employing Assumption \ref{asume_margin}.
\end{remark}

\subsubsection*{Examples}
We now demonstrate that using Theorem \ref{thm_main} can yield bounds on the misclassification excess risk in various scenarios. Further non-trivial applications are discussed in Section \ref{sc_application}.

\begin{exm}[Finite case]
	\label{exm:first:finite}
Let us begin with the special case where \(\Theta\) is a finite set, specifically, \({\rm card}(\Theta) = M < +\infty\). In this scenario, the Gibbs posterior \(\hat{\rho}_\lambda^\phi \) of~\eqref{def:Gibbspost} is a probability distribution over the finite set \(\Theta\) defined by
	$$
	\hat{\rho}_\lambda^\phi (\theta) 
	= 
	\frac{{\rm e}^{-\lambda r^\phi_n(\theta)}\pi(\theta) }{\sum_{\vartheta\in\Theta} {\rm e}^{-\lambda r^\phi_n(\vartheta)}\pi(\vartheta)}.
	$$
	As the bounds in \eqref{eq_mai_01} and \eqref{eq_mai_02} hold for all $\rho\in\mathcal{P}(\Theta) $, it holds in particular for all $\rho$ in the set of Dirac masses $ \{ \delta_\theta, \theta\in\Theta \} $. That
	\begin{equation*}
	\mathbb{E} [
\mathbb{E}_{\theta\sim \hat{\rho}_{\lambda}^\phi}
[ R_{0/1} (\theta) ] ] - R^*_{0/1}
\leq
\Psi
\inf_{\theta\in\Theta} \! 
\left\{ 
R^\phi (\theta)  - R^\phi (\theta^*) 
+ 
\frac{\overline{C}  
	\mathcal{K}(\rho\| \pi) }{n} \right\}
,
\end{equation*}
and in particular, for $ \theta = \theta^* $, this becomes
	\begin{equation*}
	\mathbb{E} [
\mathbb{E}_{\theta\sim \hat{\rho}_{\lambda}^\phi}
[ R_{0/1} (\theta) ] ] - R^*_{0/1}
\leq
\Psi
\frac{\overline{C}  
	\mathcal{K}(\delta_\theta\| \pi) }{n}
,
\end{equation*}
And,
	$
	\mathcal{K}(\delta_\theta\|\pi)
	= 
	\sum_{\theta' \in\Theta} \log\left(\frac{\delta_{\theta}(\theta' ) }{\pi(\theta' )}\right) \delta_{\theta}(\theta' ) 
	= 
	\log \frac{1}{\pi(\theta)}
	.
	$
This gives us an insight into the role of the measure \(\pi\): the bound will be tighter for \(\theta\) values where \(\pi(\theta)\) is large. However, \(\pi\) cannot be large everywhere because it is a probability distribution and that \( \sum_{\theta \in \Theta} \pi(\theta) = 1 \). The larger the set \(\Theta\), the more this total sum of 1 will be spread out, resulting in large values of \(\log(1/\pi(\theta))\). If \(\pi\) is the uniform probability distribution, then \( \log(1/\pi(\theta)) = \log(M) \), and the previous bound becomes
	\begin{equation*}
	\mathbb{E} [
\mathbb{E}_{\theta\sim \hat{\rho}_{\lambda}^\phi}
[ R_{0/1} (\theta) ] ] - R^*_{0/1}
	\leq
	\Psi \overline{C} 
	\frac{ \log(M) }{n} 
	.
	\end{equation*}
	Thus, in this case, the misclassification excess risk is of order $ \log(M)/n $.
\end{exm}

\begin{exm}
Now, we consider the continuous case where \(\Theta = \mathbb{R}^d\), the loss function is Lipschitz, and the prior \(\pi\) is a centered Gaussian: \(\mathcal{N}(0,\sigma^2 I_d)\), where \(I_d\) denotes the \(d \times d\) identity matrix. When applying Theorem \ref{thm_main}, the right-hand side in \eqref{eq_mai_01} involves an infimum over all \(\rho \in \mathcal{P}(\Theta)\). However, for simplicity and practicality, it is advantageous to consider Gaussian distributions as $ \rho = \rho_{m,s} = \mathcal{N}(m,s^2 I_d) $ with $ 	m\in\mathbb{R}^d, s>0 $. 

First, it is well known that,
$
\mathcal{K}(\rho_{m,s}\|\pi) 
= 
\frac{\|m\|^2}{2\sigma^2} 
+ \frac{d}{2} \left[
\frac{s^2}{\sigma^2} 
+ \log(\frac{\sigma^2}{s^2}) 
-1 \right]
.
$
Moreover, the risk $R^\phi $ inherits the Lipschitz property of the loss, that is, for any $(\theta,\vartheta)\in\Theta^2$, 
$ 
R^\phi (\theta) - R^\phi (\vartheta) 
\leq 
L \|\vartheta-\theta\| 
$. And, by Jensen's inequality, that $ 
\mathbb{E}_{\theta\sim\rho_{m,s} } \|\vartheta-\theta\| 
\leq $  $ 
\sqrt{\mathbb{E}_{\theta\sim\rho_{m,s} }[ \|\theta-m\|^2 ]}
$  $
\leq s \sqrt{d}.
$
Consequently, putting all thing together, with $ m= \theta^* $
\begin{equation*}
	\mathbb{E} [
\mathbb{E}_{\theta\sim \hat{\rho}_{\lambda}^\phi}
[ R_{0/1} (\theta) ] ] - R^*_{0/1}
\leq
\Psi \!\! \!\!
\inf_{	\begin{tiny}
	\begin{array}{c}
 s>0
	\end{array}	\end{tiny} }
 \!\!  \!\!
\left\{ L s \sqrt{d}
+ 
\overline{C}  
\frac{
\frac{\| \theta^* \|^2}{2\sigma^2} 
+ \frac{d}{2} \left[
\frac{s^2}{\sigma^2} 
+ \log(\frac{\sigma^2}{s^2}) 
-1 \right] }{n} \right\}
.
\end{equation*}	
Taking $ s = 1/(n\sqrt{d}) $,
\begin{align*}
	\mathbb{E} [
\mathbb{E}_{\theta\sim \hat{\rho}_{\lambda}^\phi}
[ R_{0/1} (\theta) ] ] - R^*_{0/1}
 \leq
\Psi  \left\{ \frac{L}{n}
+ 
\overline{C}   \frac{
	\frac{\| \theta^* \|^2}{2\sigma^2} 
	+ 
	\frac{d}{2} \left[
	\frac{1}{n^2 d \sigma^2} 
	+ \log( n^2 d \sigma^2 ) 
	-1 \right] }{n} \right\}
 \lesssim
\frac{d\log(n)}{n} 
.
\end{align*}
Thus, in this case, the misclassification excess risk is of order $ d\log(n)/n $. 
\end{exm}

\section{Application}
\label{sc_application}
We note that  our procedure is applicable to different classification contexts. Here, we will demonstrate it with the following two important examples.

\subsection{High dimensional sparse classifcation}

In this context, we have that $ \mathcal{X} = \mathbb{R}^d $ and that $ d> n $. Consider the class of linear classifiers, the empirical risk is now given by:
$
r_n^{0/1} (\theta) 
= 
\frac{1}{n}\sum_{i=1}^n   
\mathbbm{1} \{ Y_{i} (\theta^{\top} X_i) < 0 \} 
,
$
and the prediction risk
$
R_{0/1} (\theta) 
= 
\mathbb{E} 
\left[ 
r_n^{0/1} (\theta) 
\right]
$.
For the sake of simplicity, we put 
$
R^*  
:= 
R(\theta^*) 
$, where $ \theta^* $ is the ideal Bayes classifier. 

Our analysis is centered on a sparse setting, where we assume \( s^* < n \), with \( s^* = \| \theta^* \|_0 \), denoting the number of nonzero elements in the parameter vector. Here, we primarily focus on the hinge loss, which results in the following hinge empirical risk:
\begin{align*}
r^h_n (\theta) 
= 
\frac{1}{n}\sum_{i=1}^n ( 1 - Y_{i} \,
(\theta^\top x_i) )_+ \, ,
\end{align*}
where $ (a)_+ := \max(a,0),\forall a \in \mathbb{R} $. We consider the following Gibbs-posterior distribution:
$
\hat{\rho}^h_\lambda(\theta)
\propto
\exp[-\lambda r^h_n(\theta)] \pi(\theta)
$
where $\lambda>0$ is a tuning parameter and $\pi(\theta)$ is a prior distribution, given in \eqref{eq_priordsitrbution}, that promotes (approximately) sparsity on the parameter vector $ \theta $. Given a positive number $ C_1 $, for all $ \theta \in B_1 (C_1):= \{ \theta \in \mathbb{R}^d : \|\theta\|_1 \leq C_1 \} $, we consider the following prior,
\begin{eqnarray}
\label{eq_priordsitrbution}
\pi (\theta) 
\propto 
\prod_{i=1}^{d} 
	(\tau^2 + \theta_{i}^2)^{-2}
,
\end{eqnarray}
where $ \tau>0 $ is a tuning parameter. For technical reason, we assume that $ C_1 > 2d\tau $. This prior is known as a scaled Student distribution with 3 degree of freedom. This type of prior has been previously examined in the different sparse problems \citep{dalalyan2012mirror, dalalyan2012sparse,mai2023high}.

\begin{theorem}
	\label{thm_sparse_clasifi}
Given that \( \mathbb{E}\|X\| \leq C_{\rm x} < \infty \), Theorem \ref{theorembernstein} and Assumption \ref{asume_margin} are satisfied, and by setting \( \lambda = n/\overline{C} \), it follows that
	\begin{equation*}
	\mathbb{E} [
	\mathbb{E}_{\theta\sim \hat{\rho}_{\lambda}^h}
	[ R_{0/1} (\theta) ] ] - R^*_{0/1}
	\leq
C \frac{ s^* \log \left(d/ s^* \right)
	}{n}
	,
	\end{equation*}
	and
	\begin{equation*}
	\mathbb{E} [
	R_{0/1} ( \hat\theta) ] - R^*_{0/1}
	\leq
C \frac{ s^* \log \left(d/ s^* \right)
	}{n}
	,
	\end{equation*}	
	for some universal constant $ C>0 $ depending only on $ K,B,C_1, C_{\rm x} $.
\end{theorem}

\begin{remark}
According to Theorem \ref{thm_sparse_clasifi}, the misclassification excess rate is of order  $ s^*\log (d/ s^*)/n $ which is established as minimax-optimal in high-dimensional sparse classification, according to \cite{abramovich2018high}. This result is novel and extends the work of \cite{mai2023high}, which addresses only the misclassification excess rate in the noiseless scenario.
\end{remark}

\subsection{1-bit matrix completion}

For sake of simplicity, for any positive integer $ m $, let $ [m] $ denote $ \{1,\ldots,m\} $.

Formally, the 1-bit matrix completion problem can be defined as a classification problem as follow: we observe $(X_k,Y_k)_{k\in[n] } $ that are $n$ i.i.d pairs
from a distribution $\mathbb{P} $.
The $X_k$'s take values in $\mathcal{X} = [d_1] \times [d_2]$ and
the $Y_k $'s take values in $ \{-1,+1\} $. Hence, the $k$-th 
observation of
an entry of the matrix is $Y_k$ and the corresponding position in the
matrix is provided by $X_k=(i_k,j_k)$. 

Here, a predictor is a function \( [d_1]\times [d_2]\rightarrow\mathbb{R} \), and it can therefore be represented by a matrix \( M \). A natural approach is to employ \( M \) such that when \( (X,Y)\sim \mathbb{P} \), the predictor \( M \) predicts \( Y \) using \( {\rm sign}(M_X) \). The performance of this predictor in predicting a new matrix entry is subsequently measured by the risk
\begin{align*}
R(M) = \mathbb{E}_\mathbb{P}\left[ \mathbbm{1}(Y M_X<0)\right],
\end{align*}
and its empirical counterpart is:
$
r_n(M) = \frac{1}{n}\sum_{k=1}^n \mathbbm{1}(Y_{k} M_{X_k}<0)
= \frac{1}{n}\sum_{k=1}^n \mathbbm{1}(Y_{k} M_{i_k,j_k}<0) .
$
\\
From the classification theory~\citep{vapnik}, the best possible classifier is the Bayes classifier
\begin{align*}
\eta(x) = \mathbb{E}(Y|X=x) \quad \textrm{or equivalently} \quad \eta(i,j) = 
\mathbb{E}[Y|X=(i,j)],
\end{align*}
and equivalently we have a corresponding optimal matrix
$
M^*_{ij}
=
\textrm{sign}[\eta(i,j)] 
.
$
We define
$\overline{r_n}=r_n(M^*)$.
Note that, clearly, if two matrices $M^1$ and $M^2$ are such as, for every 
$(i,j)$, 
$\textrm{sign}(M^1_{ij})=\textrm{sign}(M^2_{ij})$ then $R(M^1)=R(M^2)$,
and obviously,
$
\forall M, \forall (i,j)\in [d_1]\times[d_2], \, 
\textrm{sign}(M_{ij})
=
M^*_{ij}  \Rightarrow  
r_n(M)=\overline{r_n}.
$

In the paper \citep{cottet20181}, the authors
deal with the hinge loss, which leads to the following so-called hinge
risk and hinge empirical risk:
\begin{align*}
R^h(M) = 
\mathbb{E}_\mathbb{P}\left[ (1-Y M_X)_+\right],
\quad
r_n^h(M) 
= \frac{1}{n}\sum_{k=1}^n (1-Y_k M_{X_k})_+.
\end{align*}
Specifically, with $ M = LR^\top $, \cite{cottet20181} define the prior distribution as the following hierarchical model:
\begin{align*}
\forall k \in [K], \quad \gamma_k &\stackrel{iid}\sim \pi^\gamma, \\
L_{i,\cdot},R_{j,\cdot}|\gamma 
&\stackrel{iid}\sim \mathcal{N}(0,\textrm{diag}(\gamma)), \forall (i,j) \in [m_1] \times [m_2], 
\end{align*}
where the prior distribution on the variances $\pi^\gamma$ is  either the
Gamma or the inverse-Gamma distribution: $\pi^\gamma=\Gamma(\alpha,\beta)$,
or $\pi^\gamma=\Gamma^{-1}(\alpha,\beta)$.

Let $\theta$ denote the parameter $\theta=(L,R,\gamma)$.
As in PAC-Bayes theory~\cite{catonibook},
the Gibbs-posterior is as follows:
\begin{align*}
\widehat{\rho}^h_\lambda(d\theta) = \frac{\exp[-\lambda r_n^h(LR^\top)]}{\int 
	\exp[-\lambda r_n^h] d\pi}\pi(d\theta)
\end{align*}
where $\lambda>0$ is a parameter to be fixed by the user.

The paper \citep{cottet20181} explores a Variational Bayes (VB) approximation, which facilitates the replacement of MCMC methods with more efficient optimization algorithms. They define a VB approximation as
$
\widetilde{\rho}_\lambda
=
\arg\min_{\rho \in 	\mathcal{F}} \mathcal{K}(\rho \| \widehat{\rho}^h_\lambda)
.
$

We define \(\mathcal{M}(r, B)\) for \( r \geq 1 \) and \( B > 0 \) as the set of pairs of matrices \((\bar{U}, \bar{V})\), with dimensions \( d_1 \times K \) and \( d_2 \times K \) respectively, that meet the conditions \(\|\bar{U}\|_{\infty} \leq B\), \(\|\bar{V}\|_{\infty} \leq B\), \(\bar{U}_{i,\ell} = 0\) for \( i > r \), and \(\bar{V}_{j,\ell} = 0\) for \( j > r \). Consistent with \cite{cottet20181, alquier2020concentration}, we assume that \( M^* = \bar{U}\bar{V}^t \) for some \((\bar{U}, \bar{V})\) in \(\mathcal{M}(r, B) \).

\begin{theorem}
	\label{thm_1bitmatrixcompletion}
	Assuming that Theorem \ref{theorembernstein} and Assumption \ref{asume_margin} holds and taking $\lambda= n/\overline{C}  $, then we have that
	\begin{equation*}
	\mathbb{E} [
	\mathbb{E}_{\theta \sim \widetilde{\rho}_{\lambda} }
	[ R_{0/1} (\theta) ] ] - R^*_{0/1}
	\leq
	C \frac{
		r(d_1 + d_2) \log(nd_1 d_2) }{n}	
	,
	\end{equation*}
	and
	\begin{equation*}
	\mathbb{E} [
	R_{0/1} ( \hat\theta) ] - R^*_{0/1}
	\leq
	C \frac{
		r(d_1 + d_2) \log(nd_1 d_2) }{n}	
	,
	\end{equation*}	
	for some universal constant $ C>0 $ depending only on $ K,B $.
\end{theorem}

\begin{remark}
The misclassification excess error rate presented in Theorem \ref{thm_1bitmatrixcompletion}, which is on the order of $ r(d_1 + d_2)/n $ (up to a logarithmic factor), is established as minimax-optimal, as demonstrated in \cite{alquier2019estimation}.
\end{remark}

\section{Concluding discussions}
\label{sc_conclusion}

This paper presents misclassification excess risk bounds for PAC-Bayesian classification, achieved through the application of a convex surrogate loss function. The methodology primarily relies on the PAC-Bayesian relative bound in expectation, coupled with the assumption of low noise condition. While our analysis assumes a bounded loss, it is worth mentioning that the findings can be extended to unbounded loss scenarios, given additional conditions as elaborated in \cite{alquier2021user}. Once the PAC-Bayesian relative bound in expectation for the chosen loss function is established, our theoretical results are applicable.

In our work, the Bernstein condition is assumed; however, it may not always be necessary. Indeed, as evidenced by several studies \cite{cottet20181,mai2023high}, in the noiseless scenario, the margin condition alone is adequate for deriving a misclassification excess risk bound. Additionally, Section 6 of \cite{alquier2019estimation} highlights that, under the hinge loss, the low-noise condition aligns with the Bernstein condition. This suggests that investigating the relationship between the Bernstein condition on convex loss and the margin condition within PAC-Bayes bounds could be a valuable area for future research.

\subsection*{Acknowledgments}
This work was supported by the Norwegian Research Council, grant number 309960, through the Centre for Geophysical Forecasting at NTNU. The author thanks Pierre Alquier for useful discussion on the Bernstein's condition.

\subsubsection*{Conflicts of interest/Competing interests}
The author declares no potential conflict of interests.

\appendix
\section{Proofs}
\label{sc_proof}

\subsection{Proof of Section \ref{sc_problem_method}}

\begin{proof}[\bf Proof os Theorem \ref{theorembernstein}]
	From Assumption \ref{assume_Lipschitz}, the loss is Lipschitz,
	\begin{equation*}
	\mathbb{E} \left\{ \left[  \phi_i(\theta) - \phi_i(\theta^*) \right]^2 \right\}
	\leq 
	L^2	\mathbb{E} \left[ \| \theta - \theta^* \|_2^2 \right]
	.
	\end{equation*}
	and from Assumption \ref{dfnbernstein},
	\begin{equation*}
	\mathbb{E} \left\{ \left[  \phi_i(\theta) - \phi_i(\theta^*) \right]^2 \right\}
	\leq 
	L^2	\mathbb{E} \left[ \| \theta - \theta^* \|_2^2 \right]
	\leq 
	L^2 K [R^\phi (\theta)-R^\phi (\theta^*) ]
	.
	\end{equation*}
	Therefore, the assumption (Definition 4.1) of
	Theorem 4.3 in \cite{alquier2021user} is satisfied with  $ L^2K $. Thus, the result is obtained by using Theorem 4.3 in \cite{alquier2021user}.
\end{proof}

\begin{proof}[\bf Proof of Theorem \ref{thm_main}]

As Assumption \ref{asume_margin} is satisfied, according to Theorem 3 in \cite{bartlett2006convexity} (taking \( \alpha = 1 \)), there exists a constant \( C > 0 \) such that
	\begin{equation*}
	\mathbb{E} [
	R_{0/1}(\theta) ] - R_{0/1}^*
	\leq
	C	\left[
	\mathbb{E}  [R^\phi(\theta) ] 
	- R^\phi (\theta^*) 
	\right]
	,
	\end{equation*}	
	integrating with respect to $ \hat{\rho}_{\lambda}^\phi $, and then using Fubini's theorem, 
\begin{align*}
	\mathbb{E} [
\mathbb{E}_{\theta\sim \hat{\rho}_{\lambda}^\phi}
[ R_{0/1} (\theta) ] ] - R^*_{0/1}
\leq
C	\left(
\mathbb{E} [ \mathbb{E}_{\theta\sim\hat{\rho}_{\lambda}^\phi} [R^\phi(\theta) ]] - R^\phi (\theta^*) 
\right)
,
\end{align*}	
	we obtain the result in \eqref{eq_mai_01}  by utilizing the result from Theorem \ref{theorembernstein}.
	
	To obtain \eqref{eq_mai_02}, as $ \phi $ is convex, an application of Jensen's inequality to Theorem \ref{theorembernstein} yields
	\begin{equation*}
	\mathbb{E} [R^\phi(\hat\theta) ] - R^\phi (\theta^*) 
	\leq 
	\mathbb{E} \mathbb{E}_{\theta\sim\hat{\rho}_{\lambda}^\phi} [R^\phi(\theta) ] - R^\phi (\theta^*) 
	\end{equation*}
	thus we can now apply Theorem 3 in \cite{bartlett2006convexity} to get that 
	\begin{equation*}
\mathbb{E} [
R_{0/1}(\hat\theta) ] - R_{0/1}^*
\leq
C	\left(
\mathbb{E}  [R^\phi(\hat\theta) ] 
- R^\phi (\theta^*) 
\right)
,
\end{equation*}
and the result is followed. This completes the proof.
\end{proof}

\begin{proof}[\bf Proof of Corollary \ref{cor_estimation}]
	As Assumptions \ref{assume_Lipschitz} and \ref{dfnbernstein} are satisfied, we obtain Theorem \ref{theorembernstein},
	\begin{equation*}
	\mathbb{E} [ \mathbb{E}_{\theta\sim\hat{\rho}_{\lambda}^\phi} [R^\phi(\theta) ]] - R^\phi (\theta^*) 
	\leq 
	2 \!\! 
	 \inf_{\rho\in\mathcal{P}(\Theta)}  
	\left\{  \mathbb{E}_{\theta\sim\rho}
	[R^\phi (\theta) ] - R^\phi (\theta^*) 
	+ 
	\frac{\overline{C}  
		\mathcal{K}(\rho\| \pi) }{n} \right\}
	.
	\end{equation*}
	Moreover, from Assumption \ref{dfnbernstein}, 
$
	\| \theta - \theta^* \|_2^2 
	\leq 
	K [R^\phi (\theta)-R^\phi (\theta^*) ]
	.
$
	Therefore, the result is obtained by combining these bounds.
\end{proof}

\subsection{Proof of Section \ref{sc_application}}

\begin{proof}[\bf Proof of Theorem \ref{thm_sparse_clasifi}]
	As the hinge loss is 1-Lipschitz, one has that
	\begin{align*}
	R^\phi (\theta) - R^\phi (\theta^*) 
	\leq
	\mathbb{E}\|X\| \|\theta - \theta^* \| 
	\end{align*}
	We define the following distribution as a translation of the prior $ \pi $,
	\begin{equation}
	\label{eq_specific_distribution}
	p_0(\beta) 
	\propto 
	\pi (\beta - \beta^*)\mathbbm{1}_{B_1(2d\tau)} (\beta - \beta^*).
	\end{equation}
	From Lemma \ref{lema_bound_prior_arnak}, we have, for $ \rho:= p_0 $, that
	\begin{align*}
	\int [ R^\phi (\theta) - R^\phi (\theta^*) ] p_0 (d \theta)
		\leq
C_{\rm x}	\int \| \beta- \beta^* \| p_0(d \beta)
	\leq
C_{\rm x} \left(	\int \| \beta- \beta^* \|^2 p_0(d \beta) \right)^{1/2}
 \leq
C_{\rm x} \sqrt{ 4d\tau^2 }
	\end{align*}
	and
	$$
	\mathcal{K}(p_0 \| \pi)
	\leq
	4 s^* \log \left(\frac{C_1 }{\tau s^*}\right)
	+
	\log(2)
	.
	$$
	Plug-in these bounds into inequality \eqref{eq_mai_01}, one gets that
	\begin{equation*}
	\mathbb{E} [
	\mathbb{E}_{\theta\sim \hat{\rho}_{\lambda}^\phi}
	[ R_{0/1} (\theta) ] ] - R^*_{0/1}
	\leq
	\Psi \!\!\!
	\inf_{\tau \in (0,C_1/2d)} 
	\!
	\left\{  
C_{\rm x} 2 \tau \sqrt{ d }
	+ 
	\frac{C_1 
		4 s^* \log \left(\frac{C_1 }{\tau s^*}\right)
		+
		\log(2)
	}{n} \right\}
	,
	\end{equation*}
	and the choice $ \tau = ( C_{\rm x} n\sqrt{d})^{-1} $ leads to
	\begin{align*}
	\mathbb{E} [
	\mathbb{E}_{\theta\sim \hat{\rho}_{\lambda}^\phi}
	[ R_{0/1} (\theta) ] ] - R^*_{0/1}
	\leq
	\Psi 
	\left\{  \frac{2}{n} 
	+ 
	\frac{C_1 4 s^* \log \left(\frac{ C_{\rm x} C_1 n\sqrt{d}}{ s^*}\right)
		+
		\log(2)	}{n} \right\}
	\leq c 
	\frac{ s^* \log \left( d/ s^*\right)
	}{n}
	,
	\end{align*}
	for some positive constant $ c $ depending only on $ L,K,B, C_1, C_{\rm x} $.
	A similar argument application to inequality \eqref{eq_mai_02}, one gets that
	\begin{align*}
	\mathbb{E} [
	R_{0/1} ( \hat\theta) ] - R^*_{0/1}
	\lesssim 
	\frac{ s^* \log \left(d/ s^*\right)
	}{n}
	.
	\end{align*}
	The proof is completed.
\end{proof}

\begin{proof}[\bf Proof of Theorem \ref{thm_1bitmatrixcompletion}]
	Using similar argument as in the proof of Theorem 4.3 in \cite{alquier2021user} (see also the proof of Theorem 4.3 in \cite{alquier2016properties}), one obtains that
	\begin{equation*}
	\mathbb{E} [ \mathbb{E}_{\theta \sim \widetilde{\rho}_{\lambda}} [R^\phi(\theta) ]] - R^\phi (\theta^*) 
	\leq 
	2  \inf_{\rho\in\mathcal{F} }  
	\left\{  \mathbb{E}_{\theta\sim\rho}
	[R^\phi (\theta) ] - R^\phi (\theta^*) 
	+ 
	\frac{C_1 
		\mathcal{K}(\rho\| \pi) }{n} \right\}
	.
	\end{equation*}	 
	A similar argument as in Theorem \ref{thm_main}, 
	\begin{equation*}
	\mathbb{E} [ \mathbb{E}_{\theta \sim \widetilde{\rho}_{\lambda}} [R_{0/1} (\theta) ]] - R_{0/1}  (\theta^*) 
	\leq 
	\Psi
	\inf_{\rho\in\mathcal{F} }  
	\left\{  \mathbb{E}_{\theta\sim\rho}
	[R^\phi (\theta) ] - R^\phi (\theta^*) 
	+ 
	\frac{C_1 
		\mathcal{K}(\rho\| \pi) }{n} \right\}
	.
	\end{equation*}	
	As the hinge loss is 1-Lipschitz, and noting that $ \theta^* = M^*, \theta = LR^\top
	$,one has that
	\begin{align*}
	R^\phi (\theta) - R^\phi (\theta^*) 
	\leq
	\|\theta - \theta^* \|
	= 
	\|LR^\top - M^* \|
	\end{align*}	
	Given \( B > 0 \) and \( r \geq 1 \), for any pair \((\bar{U}, \bar{V}) \in \mathcal{M}(r, B)\), we define 
	\begin{equation}
	\label{eq_specific_rhon}
	\rho_n({\rm d}U,{\rm d}V,{\rm d}\gamma) 
	\propto 
	\mathbf{1}_{(\|U-\bar{U}\|_{\infty} \leq \delta,\|U-\bar{U}\|_{\infty} \leq \delta)} \pi({\rm d}U,{\rm d}V,{\rm d}\gamma),
	\end{equation}
	where \(\delta \in (0, B)\) to be selected later.
	For any $(U,V)$ in the support of $\rho_n$, given in \eqref{eq_specific_rhon}, one has that
	\begin{align*}
	\|M^* - UV^t\|_F 
	& = 
	\| \bar{U}\bar{V}^t-\bar{U}V^t+\bar{U}V^t-UV^t\|_F \\
	& \leq  \|\bar{U}(\bar{V}^t-V^t)\|_F+\|(\bar{U}-U)V^t\|_F \\
	& \leq  \|\bar{U}\|_F \|\bar{V}-V\|_F+\|\bar{U}-U\|_F \|V^t\|_F \\
	& \leq  
	d_1 d_2 \|\bar{U}\|_{\infty}^{1/2} \|\bar{V}-V\|_\infty^{1/2}
	+d_1 d_2 \|V\|_{\infty}^{1/2} \|\bar{U}-U\|_\infty^{1/2} \\
	& \leq 
	d_1 d_2 \delta^{1/2}  
	[B^{1/2} + (B+\delta)^{1/2}  ] 
	\\
	& \leq 
	2 d_1 d_2 \delta^{1/2} (B+\delta)^{1/2} 
 \leq 
	2^{3/2} d_1 d_2 \delta^{1/2} B^{1/2}  
	.
	\end{align*}
	Thus, with $\delta=B/[8(nd_1 d_2)^2]$, 
	one gets that
	$$  
	\mathbb{E}_{\theta\sim\rho_n}
	[R^\phi (\theta) ] - R^\phi (\theta^*)  
	\leq B/n
	.
	$$
	Now, from Lemma \ref{lm_boundKL}  with $\delta=B/[8(nd_1 d_2)^2]$, we have that
	\begin{equation*}
	\frac{1}{n}
	\mathcal{K}(\rho_n \| \pi) 
	\leq 
	\frac{
		2(1+2a) r(d_1 + d_2) \left[ \log(nd_1 d_2)
		+
		C_a
		\right] }{n}
	.
	\end{equation*}	
	Putting all together, 
	\begin{align*}
	 \mathbb{E} [ \mathbb{E}_{\theta \sim \widetilde{\rho}_{\lambda}} [R_{0/1} (\theta) ]] - R_{0/1}  (\theta^*) 
	&\leq 
	C
	\left\{ \frac{ B}{n}
	+ 
	\frac{
		2(1+2a) r(d_1 + d_2) \left[ \log(nd_1 d_2)
		+
		C_a	\right] }{n}
	\right\}
	\\
	&
	\lesssim 
	\frac{
		r(d_1 + d_2) \log(nd_1 d_2) }{n}	
	,
	\end{align*}	
	for some numerical constant $ C>0 $ depending only on $ a, C_1 $. 
	The proof is completed.
\end{proof}

\begin{lemma}
	\label{lema_bound_prior_arnak}
	Let $p_0 $ be the probability measure defined by (\ref{eq_specific_distribution}). If
	$d\geq 2$ then
	$
	\int_\Lambda \| \beta- \beta^* \|^2 p_0(d \beta)
	\leq
	4d\tau^2 
	,
	$
	and
	$
	\mathcal{K}(p_0 \| \pi)
	\leq
	4 s^* \log \left(\frac{C_1 }{\tau s^*}\right)
	+
	\log(2)
	.
	$
\end{lemma}
\begin{proof}
	The proof can be found in \cite{mai2023high}, which utilizes results from \cite{dalalyan2012mirror}.
\end{proof}

\begin{lemma}
	\label{lm_boundKL}
	Put $ C_a:= \log(8\sqrt{\pi}\Gamma(a)2^{10a+1})+3 $
	and with $\delta=B/[8(nd_1 d_2)^2]$ that satisfies $0<\delta<B$, we have for $ \rho_n $ in \eqref{eq_specific_rhon} that
$
	\mathcal{K}(\rho_n \| \pi) \leq 2(1+2a) r(d_1 + d_2) \left[ \log(nd_1 d_2)
	+
	C_a
	\right].
$
\end{lemma}
\begin{proof}
	This result can found in the proof of Theorem 4.1 in \cite{alquier2020concentration}.	
\end{proof}

%	\bibliographystyle{apalike}
%	\bibliography{refs_sparselogistic}

\end{document}